\documentclass[sigconf]{acmart}

\def\BibTeX{{\rm B\kern-.05em{\sc i\kern-.025em b}\kern-.08emT\kern-.1667em\lower.7ex\hbox{E}\kern-.125emX}}
    
\copyrightyear{2020}
\acmYear{2020}
\setcopyright{acmlicensed}

\usepackage{amsmath,amsfonts,bm}

\def\eqref#1{equation~\ref{#1}}

\def\1{\bm{1}}

\def\vh{{\bm{h}}}

\def\vm{{\bm{m}}}

\DeclareMathAlphabet{\mathsfit}{\encodingdefault}{\sfdefault}{m}{sl}
\SetMathAlphabet{\mathsfit}{bold}{\encodingdefault}{\sfdefault}{bx}{n}

\def\gF{{\mathcal{F}}}
\def\gG{{\mathcal{G}}}

\def\gN{{\mathcal{N}}}

\usepackage{subcaption}
\usepackage{booktabs}

\begin{document}

\title[Examining COVID-19 Forecasting using Spatio-Temporal GNNs]{Examining COVID-19 Forecasting using Spatio-Temporal Graph Neural Networks}

\author{Amol Kapoor*, Xue Ben*, Luyang Liu, Bryan Perozzi, Matt Barnes, Martin Blais, Shawn O'Banion}
\email{[ajkapoor, sherryben, luyangliu, hubris, mattbarnes, blais, obanion]@google.com}
\affiliation{%
	\institution{Google Research}
}
\thanks{* Both authors contributed equally to this work}
\renewcommand{\shortauthors}{Kapoor, Ben et al.}

\begin{abstract}

In this work, we 
examine a novel forecasting approach for \mbox{COVID-19} case prediction that uses Graph Neural Networks and mobility data. In contrast to existing time series forecasting models, the proposed approach learns from a single large-scale spatio-temporal graph, where nodes represent the region-level human mobility, spatial edges represent the human mobility based inter-region connectivity, and temporal edges represent node features through time. We evaluate this approach on the US county level \mbox{COVID-19} dataset, and demonstrate that the rich spatial and temporal information leveraged by the graph neural network allows the model to learn complex dynamics. We show a 6\% reduction of RMSLE and an absolute Pearson Correlation improvement from 0.9978 to 0.998 compared to the best performing baseline models.
This novel source of information combined with graph based deep learning approaches can be a powerful tool to understand the spread and evolution of \mbox{COVID-19}. We encourage others to further develop a novel modeling paradigm for infectious disease based on GNNs and high resolution mobility data. 

\end{abstract}

\maketitle

\section{Introduction}

From late 2019 to early 2020, \mbox{COVID-19} went from a local outbreak to a worldwide pandemic, one that has infected over 6.67M people and resulted in over 391K deaths worldwide~\cite{whoDashboard}. Between large-scale country-wide quarantines and `lockdowns', \mbox{COVID-19} is responsible for an estimated 3-10 trillion dollars in economic damage to the global economy~\cite{UNCovidEcoEstimate}. In a state of pandemic, the ability to accurately forecast caseload is extremely important to help inform policymakers on how to provision limited healthcare resources, rapidly control outbreaks, and ensure the safety of the general public.

In order to prepare, understand, and control the spread of the disease, researchers worldwide have come together in a collaborative effort to model and forecast \mbox{COVID-19}. Based on our review of the literature, there are two popular approaches for such epidemiological modelling. One is the mechanistic approach -- for example, compartmental and agent based models that hard-code predefined disease transmission dynamics at either the population level~\cite{yang2020JTD, PeiShaman2020} or the individual level~\cite{chang2020modelling}. The other is the time series learning approach -- for example, applying curve-fitting~\cite{murray2020forecasting}, Autoregression (AR) ~\cite{DurbinKoopman2012}, or deep learning~\cite{yang2020JTD} on time series data. 

These approaches often assume a relatively closed-system, where forecasts for a given location are dependent only on information from that location or some observed patterns from other locations. In practice, we intuit that infection data on inter-regional interactions provides a unique and highly meaningful avenue for modelling forecasts. In other words, it is reasonable that a region’s future disease cases are dependent on its own historical information as well as other regions', people traveling to/out of this region and regions with similar epidemic patterns, etc. Based on this insight, we believe we can improve forecast accuracy by 1) utilizing more accurate real-time data that can describe the inter-region interactions and region-level mobility and 2) developing a unifying approach that can encompass both the temporal and spatial interactions for infectious disease modeling. Historically, this kind of regional movement is difficult to capture. However, researchers have correctly noticed that the widespread use of GPS enabled mobile devices provides a novel and highly accurate source of mobility data, and have called upon the epidemiological community to make ample use of this powerful new data source~\cite{oliver2020mobile, buckee2020aggregated}.

In this work, we focus on the problem of forecasting \mbox{COVID-19} at the county level in the United States.
We propose a spatio-temporal graph neural network that can learn the complex dynamics inherent to disease modeling, and use this model to make forecasts on \mbox{COVID-19} daily new cases from fine-grained mobility data. 
We run several experiments showing the power of novel mobility data within the GNN framework, and conclude with an analysis of mobility data and its potential in tracking disease spread.

\section{Background}

\subsection{Mobility Data in Graphs}
Obtaining fine-grained human mobility data that can effectively capture the inter- and intra-region flows of human activity has become significantly more feasible in the last decade. In addition to being vital for accurately modeling disease spread, these data sources are especially important to understand the efficacy of non-pharmaceutical interventions (NPI) against \mbox{COVID-19}, such as social distancing, shelter-in-place, and the shut-down of interstate and international travel.

The rapid work of the epidemiological academic community was vital for understanding the role of international flights in the early spread of \mbox{COVID-19} to different countries~\cite{Adiga2020, yang2020JTD}, while epidemic curve fitting analysis for \mbox{COVID-19} on the SafeGraph dataset~\cite{woody2020projections} helped to better model the effects and efficacy of social distancing. We build on those efforts by examining and utilizing two Google mobility datasets, which offer a global and comprehensive view of inter- and intra-region human mobility. These datasets are described in more detail in \autoref{sec:data}.

\subsection{Spatio-Temporal Graph Neural Networks}
Graphs are natural representations for a wide variety of real-life data in social, biological, financial, and many other fields. Recently, graph neural network (GNN) based deep learning methods~\cite{zhou2018graph,wu2019comprehensive,zhang2018Deep, battaglia2018relational, bronstein2017geometric} have shown superior performance on several tasks, including semi-supervised node classification~\cite{kipf2016semi, hamilton2017inductive, velickovic2017graph}, link prediction ~\cite{kipf2016variational,bojchevski2018deep,zhang2018link}, community detection~\cite{chen2018supervised,shaham2018spectralnet, kawamoto2018MeanfieldTO}, graph classification~\cite{gilmer2017neural, xu2018powerful, niepert2016learning}, and recommendations~\cite{monti2017GeometricMC, ying2018graph}.

Spatio-temporal graphs are a kind of graph that model connections between nodes as a function of time and space, and have found uses in a wide variety of fields~\cite{reinhart2018review}. GNNs have been successfully applied to spatio-temporal traffic graphs~\cite{diao2019dynamic} and (especially relevant to this work) spatio-temporal influenza forecasting~\cite{deng2019graph}. In these latter two cases, temporal dependencies were primarily incorporated at the model level, either through decomposition of a dynamic Laplacian matrix or through a recurrent neural net.

\section{Method}
\subsection{Graph Neural Networks}
\label{sec:background_gnn}
The core insight behind graph neural network models is that the transformation of the input node's signal can be coupled with the propagation of information from a node's neighbors in order to better inform the future hidden state of the original input. This is most evident in the message-passing framework proposed by~\citet{gilmer2017neural}, which unifies many previously proposed methods. In such approaches, the update at layer $(l+1)$ is:
\begin{align}
\label{eq:message_passing}
\vm_i^{(l+1)} = \sum_{j\in\gN(i)} \gF^{(l)}\big(\vh_i^{(l)}, \vh_j^{(l)}\big)  ,\quad
\vh_i^{(l+1)} = \gG^{(l)} \big(\vh_i^{(l)}, \vm_i^{(l+1)}\big)
\end{align}
where $\gF^{(l)}$ and $\gG^{(l)}$ are learned message functions and node update functions respectively, $\vm^{(l)}$ are the messages passed between nodes, and $\vh_i^{(l)}$ are the node representations. The computation is carried out in two phases: first, messages are propagated along the neighbors; and second, the messages are aggregated to obtain the updated representations.

\subsection{Modelling the COVID-19 Graph}
In infectious disease modeling, we usually have multiple time-series sequences that represent the observables of transmission dynamics in each location. The prediction problem is usually formulated as a regression learning task that takes in a certain time series $t-k,\dotsc,t-1, t$ and outputs a single value $t+1$ or future time series $t+1, t+2, \dotsc$ as forecasted values. However, time series make a poor fit for modeling human mobility across locations. Mobility data is naturally represented as a spatial-graph, where any individual node represents a location $i$ that is connected to an arbitrary number of other nodes $j, l, m, \dotsc$, and where edge-weights correspond to measures of human mobility between the nodes.

\begin{figure}[t!]
	\centering
	\includegraphics[width=0.6\linewidth]{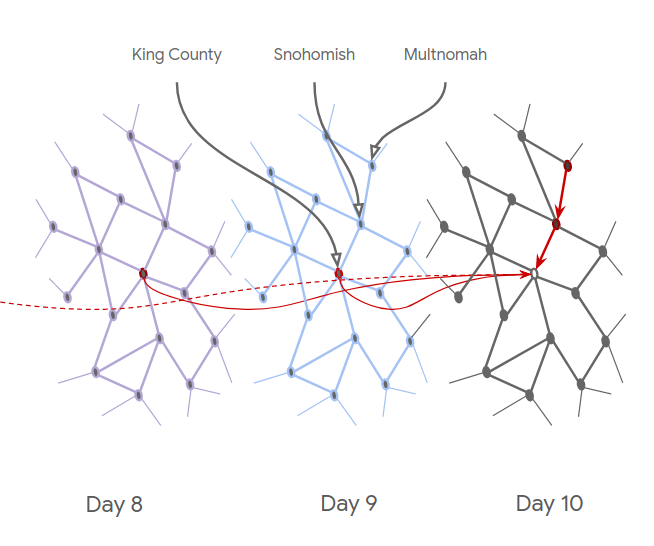}
	\caption{A slice of the COVID-19 graph showing spatial and temporal edges (highlighted in red) across three days. Each slice represents spatial connections between counties, while the connections between slices represent temporal relationships. For clarity, only temporal edges to the center node are shown; in practice, every node in the graph has direct temporal edges to nodes in $d$ previous days.}
	\label{fig:graph}
\end{figure}

In order to model spatial and temporal dependencies, we create a graph with different edge types. In the spatial domain, edges represent direct location-to-location movement and are weighted based on mobility flows normalized against the intra-flow (in other words, the amount of flow internal to the location). In the temporal domain, edges simply represent binary connections to past days. The graph manifests as ~100 stacked layers. Each layer represents the county connectivity graph for that day, with the bottom layer representing Feb 22nd, 2020 (when cases began appearing in earnest in the US), and the top layer representing May 31st, 2020. Each node within each layer has direct edges to the 7 nodes directly before it in time, i.e. a week's worth of temporal information. We provide a visual of a part of the graph in \autoref{fig:graph}. 

\subsection{Skip-Connections Model}

\begin{figure}[t!]
	\centering
	\includegraphics[width=0.6\linewidth]{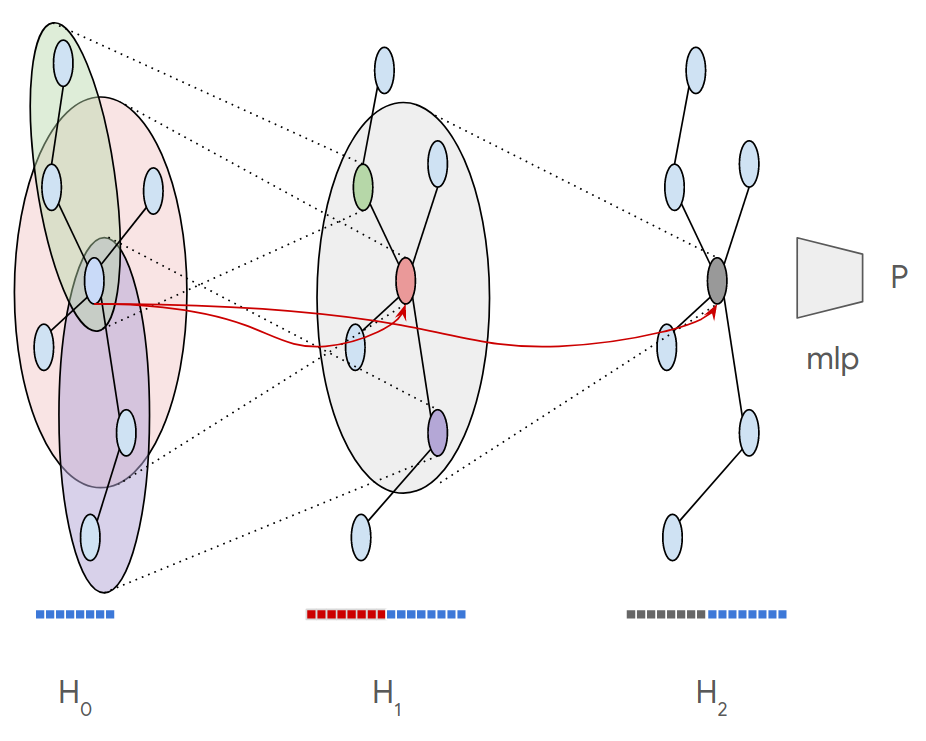}
	\caption{A visualization of a 2-hop Skip-Connection model. Multiple layers of spatial aggregations are used on temporal embedding vectors. At each layer, the embedding of the seed-node (represented in blue) is concatenated and propagated up to the next embedding layer. The final embedding is passed through an MLP and used to predict $P$.}
	\label{fig:skiphop}
\end{figure}

For our graph convolutions, we use a version of the spectral graph convolution model proposed by~\citet{kipf2016semi}, modified with skip-connections between layers to avoid diluting the self-node feature state. Specifically, the output of each layer is concatenated with a learned embedding from the temporal node features. The model prediction $P$ can be represented as:
\begin{align}
\mathbf{H_0} &= \textrm{mlp}(\mathbf{x_t | x_{t-1} | ... | x_{t-d}}) \\
\mathbf{H_{l+1}} &= \sigma(\mathbf{\hat{A}} \mathbf{H_l} \mathbf{W_l}) \enskip |\enskip \mathbf{H_0} \\
\mathbf{P} &= \textrm{mlp}(\mathbf{H_s})
\end{align}
where $H$ represents the hidden state at layer $l$, $\hat{A}$ is the spectral normalized adjacency matrix, $W$ is the learned weight matrix at layer $l$, $|$ is the concat operator, and $\sigma$ is a nonlinearity (in our case, a relu). See \autoref{fig:skiphop} for a visual representation. The first embedding, $H_0$, is simply the output of an mlp over the node's temporal features $x$ at time $t$ reaching back $d$ days, while the final prediction is the output of an mlp over $s$ spatial hops. 

\section{Experiments}

\subsection{Data}
\label{sec:data}
We make use of three datasets: the New York Times (NYT) \mbox{COVID-19} dataset\footnote{https://github.com/nytimes/covid-19-data}, the Google COVID-19 Aggregated Mobility Research Dataset, and the Google Community Mobility Reports\footnote{https://www.google.com/covid19/mobility/}. The Aggregated Mobility Research Dataset helps us understand the quantity of movement, while the Community Mobility Reports helps us understand the dynamics of various types of movement. Together, these datasets add significant lift to the standard node features provided by the NYT. 

\subsubsection{Common Node Features}
Each node contains features for state, county, day, past cases, and past deaths. The latter two are represented as normalized vectors that stretch back $d$ days. We use \mbox{COVID-19} case and death count numbers published by the New York Times~\cite{nytimes}, which includes daily reports of new infections and deaths at both state and county level in US.  

\subsubsection{Aggregated Mobility Research Dataset}
\begin{figure*}[t]
    \vspace{-1em}
    \centering
    \begin{minipage}[t]{0.3\textwidth}
    \includegraphics[width=0.95\textwidth]{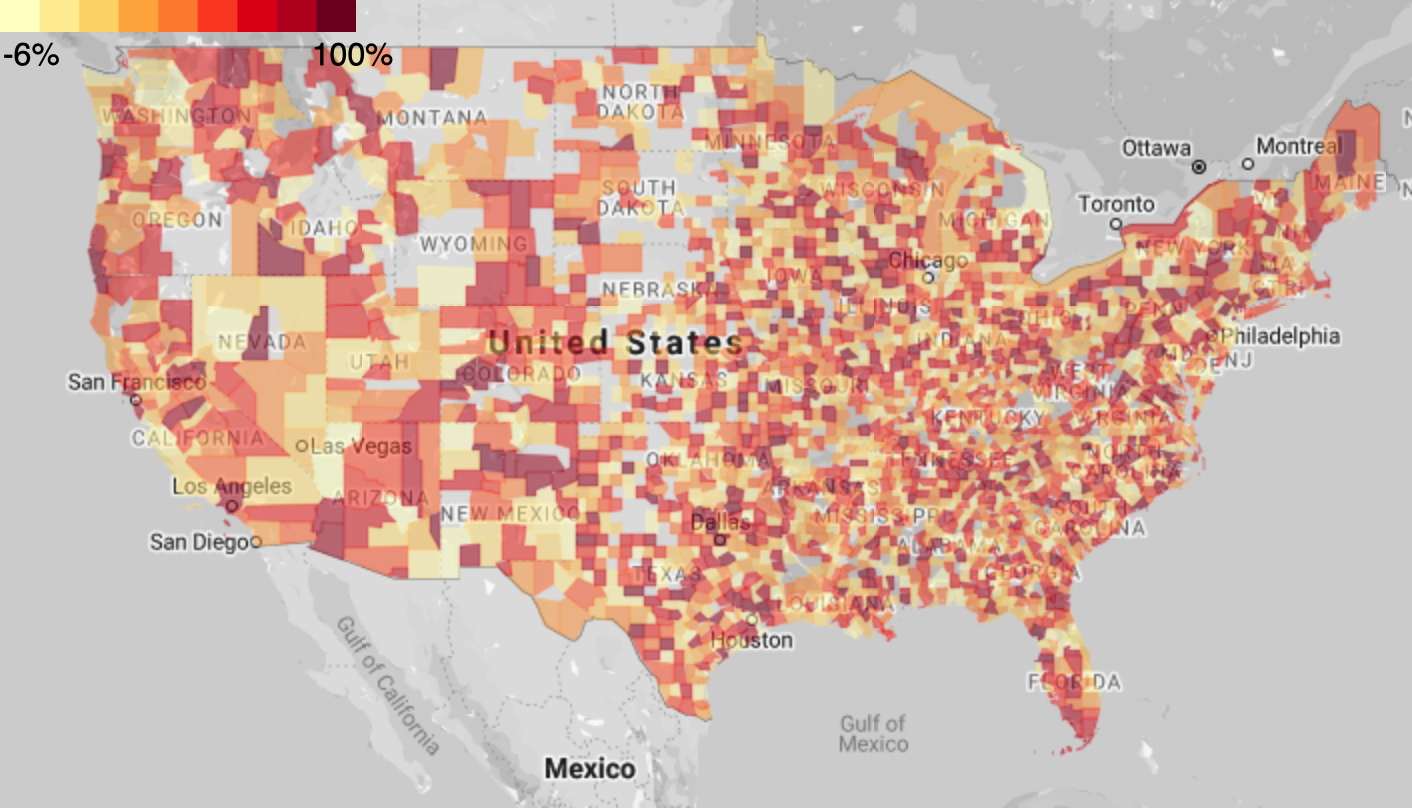}
    \caption{The reduction of inter-county mobility flow for US counties, comparing flows in April to baseline values in the first 6 weeks of 2020.}
    \label{fig:us_county_inter_flow_redu}
    \end{minipage}\hfill
    \begin{minipage}[t]{0.3\textwidth}
    \includegraphics[width=0.95\textwidth]{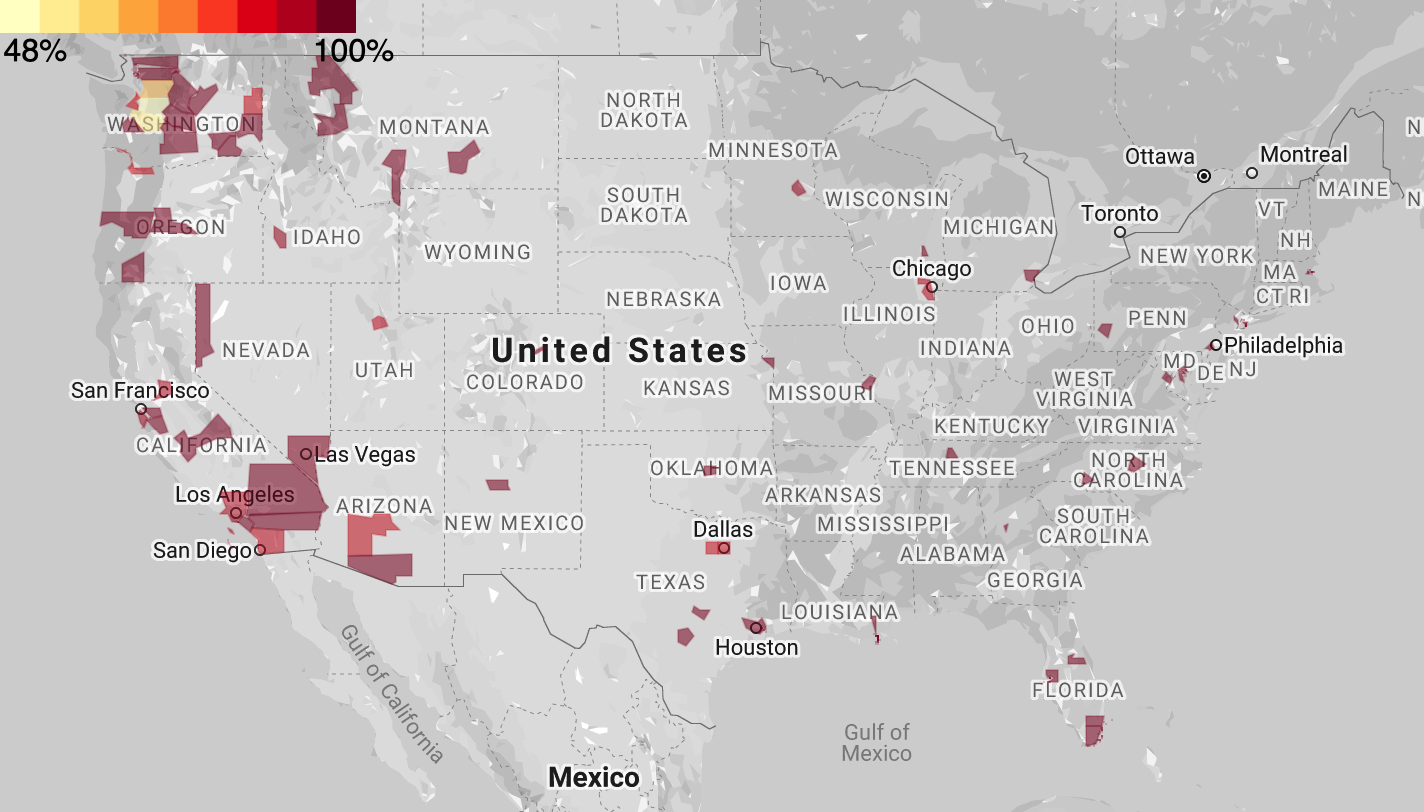}
    \caption{The reduction of inflow for King county from various US counties. Note that because King county has an airport, it has direct edges to US counties that may be physically distant.}
    \label{fig:king_county_inter_inflow_redu}
    \end{minipage}\hfill
    \begin{minipage}[t]{0.3\textwidth}
    \includegraphics[width=0.95\textwidth]{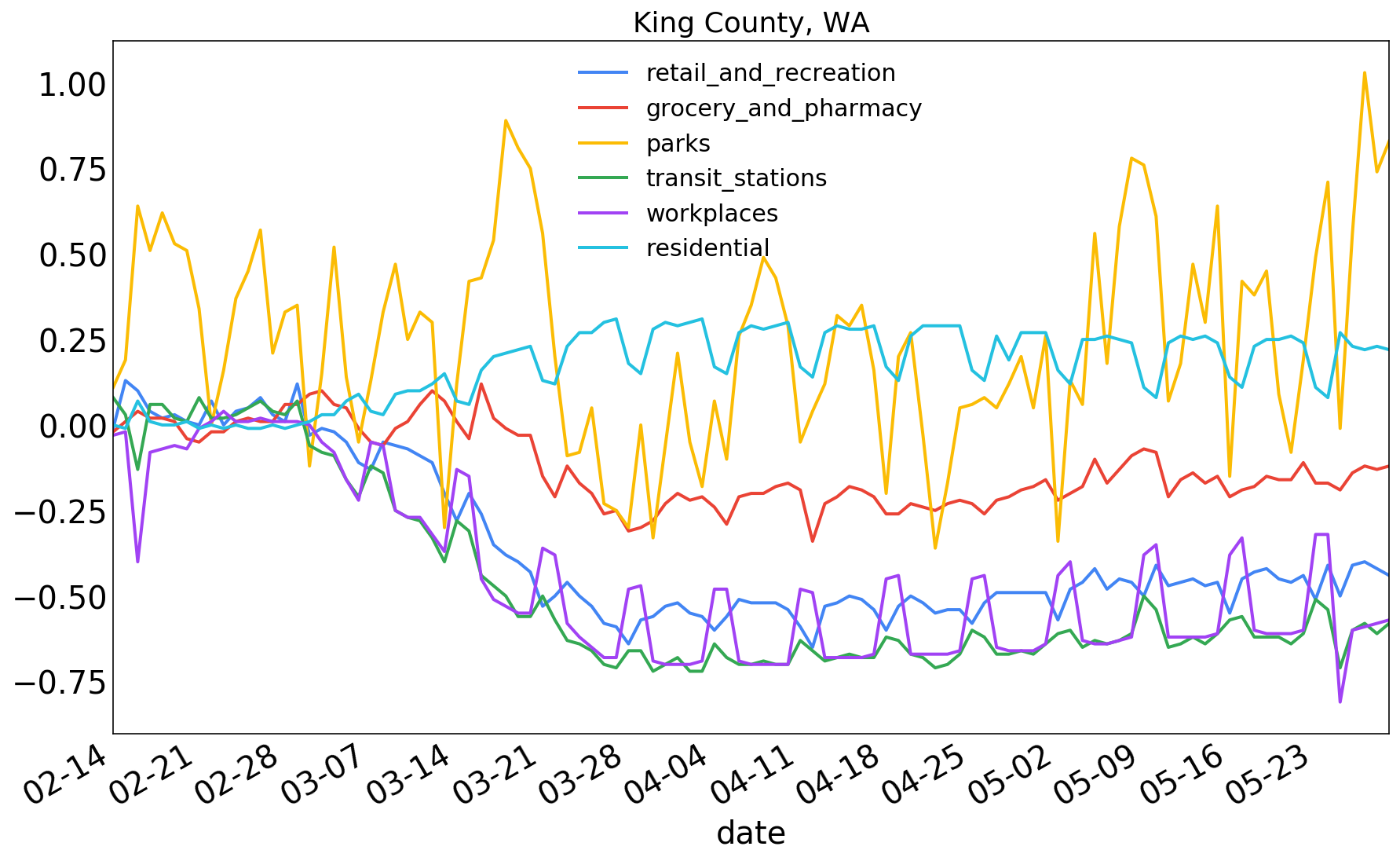}
    \caption{The mobility trends for King county. There are dramatic reductions in many of the mobility categories in late March due to non-pharmaceutical interventions like social distancing and quarantine.}
    \label{fig:king_pv_mob_trends}
    \end{minipage}\hfill
\end{figure*}

The Google COVID-19 Aggregated Mobility Research Dataset aggregates weekly flows of users from region to region, where the region is at a resolution of 5km$^2$. The flows can be further aggregated to obtain inter-county flows and intra-county flows(source and destination regions are in the same county) to build our proposed graph network. This information is useful for understanding how people move before and during the pandemic -- for example, \autoref{fig:us_county_inter_flow_redu} shows the reduction in inter-county flows in US counties in April, compared to a January baseline. \autoref{fig:king_county_inter_inflow_redu} illustrates the change in mobility to King County, Washington, where mobility dropped by nearly $100\%$ from distant counties, likely due to reductions in air travel. By comparison, reductions are less strong from nearby counties, e.g. $64\%$ reduction from Snohomish County, Washington. For a full description of how the Aggregated Mobility Research Dataset is created, see (Appendix) \ref{agg-epi-desc}.

\subsubsection{Community Mobility Reports}
The Community Mobility Reports summarize mobility trends at various categories of places that are aggregated at the county level. The categories include: grocery and pharmacy, parks, transit stations, workplaces, residential, and retail and recreation. The dataset was normalized to have 0 as the `normal' mobility based on median value for the corresponding day of the week, during the 5-week period Jan 3$-$Feb 6, 2020~\cite{googleMobReport2020}, and deviations are measured as the relative changes in mobility from the baseline. A value of -0.25 under transit stations therefore represents a 25\% reduction in visits to public transit stations compared against baseline. \autoref{fig:king_pv_mob_trends} provides a visual example of the daily mobility changes in King County, Washington for each category in Google's Community Mobility Reports. 

\subsubsection{Limitations of Data Sources}
These results should be interpreted in light of several important limitations. First, the Google mobility data is limited to smartphone users who have opted in to Google's Location History feature, which is off by default. These data may not be representative of the population as whole, and furthermore their representativeness may vary by location. Importantly, these limited data are only viewed through the lens of differential privacy algorithms, specifically designed to protect user anonymity and obscure fine detail. Moreover, comparisons across rather than within locations are only descriptive since these regions can differ in substantial ways. This data can be viewed as similar to the data used to show how busy certain types of places are in Google Maps --- for example, helping identify when a local business tends to be the most crowded. 

We also note that there are significant other factors not captured in any of these datasets, such as the increased prevalence of wearing masks or changes in the weather. These factors, combined with increased awareness, can effectively reduce the transmission even when mobility remains unchanged. We encourage future work that explores the addition of these external features.

\subsection{Hyperparameters, Architectures, and Splits}

Unless explicitly stated otherwise, for all of our GNN experiments, we use a 7 day (i.e. one week) time horizon and look over 2 hops of spatial data (using the 32 neighbors with the highest edge weight for each hop). GNN models were implemented in Tensorflow. We utilize an ADAM optimizer with learning rate set to 1e-5. We use a two hop spatial model with a single layer MLP on either side. Therefore, we have four hidden layers -- an initial embedding layer, the two hops of spatial aggregation, and the final prediction layer. The hidden layer architecture for $W_0$, $W_1$, $W_2$, and $W_3$ are [64, 32, 32, 32], respectively. Each layer has a dropout rate of 0.5, and a l2 regularization term of 5e-4. GNN models were trained for 1M steps with a MSLE regression loss. 

All models were trained to predict the change in the number of cases on day $t+1$, given previous information. We have data from January 1st onwards; however, we do not observe cases in the US until late February. As a result, we use data from days 59-120 (roughly, March and April, 2020) for training, and data from days 120 to 150 (roughly, May, 2020) was used for testing. For each model, we look at the top 20 counties by population. The reported  values are averaged across all counties for all thirty days of inference.

\subsection{Baselines}

To evaluate the benefits of the GNN framework, we compare against a range of popular methods as baselines. For all of our baselines, we examine how region-level mobility features, such as aggregated flows and place visit trends, affect our results. `No Mob' versions of our baselines indicate that these baselines do not utilize any mobility information.

\subsubsection{Previous Day}
Next day case prediction is highly correlated with features from the previous day. We use two previous day baselines. For Previous Delta, we predict that the delta in the number of cases will be the same as the delta from the previous day. For Previous Cases, we predict that the delta in the number of cases will be 0 (and that the actual number of cases will be the same as the previous day). These baselines help us understand what lift, if any, our models are able to extract from the rest of the provided features; however, we do not treat these as `model` baselines in our analysis.

\subsubsection{ARIMA}
We utilize a univariate ARIMA model that treats the time dependent daily new cases as a univariate time series that follows a fixed dynamic. Each day's new case count is dependent on the previous $p$ days of observations and the previous $q$ days of estimation errors. We selected the order of the ARIMA model using Akaike Information Criterion (AIC) and Bayesian Information Criterion (BIC) to balance model complexity and generalization, we minimize parameters by using a constant trend with $\mathrm{ARIMA}(7,1,3)$.

\subsubsection{LSTM and Seq2Seq}

Our LSTM baseline contains a stack of two LSTM layers (with 32, 16 units respectively) and a final dense layer. The LSTM layers encode sequential information from input through the recurrent network. The dense connected layer takes the final output from the second LSTM layer and outputs a vector of size four, which is equal to the number of steps ahead predictions needed.

The Seq2Seq model has an encoder-decoder architecture, where the encoder is composed of a fully connected dense layer and a GRU layer that can learn from sequential input and return a sequence of encoded outputs in a final hidden state. The decoder is an inverse of the encoder. The dense layer is 16 units and the GRU layer is 32 units. To match common practice, we apply Bahdanau attention~\cite{bahdanau14ICLR} on the sequence of encoder outputs at each decoding step to make next step prediction. 
Both  the LSTM and Seq2Seq models, we use a Huber loss, an Adam optimizer with a learning rate of 0.02, and a dropout rate of 0.2 for training. During inference, both models observe data from the previous 10 days in order to make a prediction about the next day in the sequence.

\subsection{Case Prediction Performance}

In \autoref{tab:graph_vs_baselines_comparison}, we compare the forecasting performance of the spatio-temporal GNN with a range of baseline models. We report the RMSLE and Pearson Correlation for the predicted caseload (RMSLE, Corr), calculated as the sum of the predicted delta and the previous day's cases. We aggregate the performance metrics from top 20 populated counties in US. We note that we can trivially achieve a high correlation because the problem framing naturally relies on the general trend of the data from time $t$ -- in fact, the Previous Cases baseline achieves the highest case correlation overall. To account for this, we also report the RMSLE and Pearson Correlations for the case deltas ($\Delta$ RMSLE, $\Delta$ Corr), even though we expect the ground truth values to be confounded by unaccounted variables like the availability of testing centers or whether it is a workday.

We find that the GNN successfully outperforms our baselines, achieving either best or second-best score on each evaluation metric. Further, we note that for all of our deep models, introducing additional mobility data improves results. Interestingly, introducing mobility data resulted in worse performance for the ARIMA baseline. ARIMA assumes fixed dynamics and a linear dependence on the county-level mobility -- while this helps the ARIMA model in the early stages of the epidemic, when there was a strong positive correlation between reduced mobility and daily new cases, it may cause the model to under-perform with the increase of mobility in late May.

\begin{table}
\begin{tabular}{l|cccc}
\toprule 
Model & RMSLE & Corr & $\Delta$ RMSLE & $\Delta$ Corr  \\
\midrule
Previous Cases & 0.0226 & \textbf{0.9981} & 4.7879 & NaN \\
Previous Delta & 0.0127 & 0.9965 & 0.9697 & 0.1854 \\
\midrule
No Mob ARIMA & 0.0124 & 0.9968 & 0.9217 & 0.1449 \\
ARIMA & 0.0144 & 0.9952 & 0.9624 & 0.0966 \\
No Mob LSTM & 0.0125 & 0.9978 & 0.9172 & 0.1540 \\
LSTM & 0.0121 & 0.9978 & 0.9163 & 0.1863 \\
No Mob Seq2Seq & 0.0118 & 0.9976 & \underline{0.8467} & 0.1020 \\
Seq2Seq & \underline{0.0116} & 0.9977 & 0.8634 & \textbf{0.2802} \\
GNN & \textbf{0.0109} & \underline{0.9980} & \textbf{0.7983} & \underline{0.2230} \\
\bottomrule
\end{tabular}
\caption{Summary of model performances.}
\label{tab:graph_vs_baselines_comparison}
\end{table}
\section{Conclusion}
In this work we developed a graph neural network based approach for \mbox{COVID-19} forecasting with spatio-temporal mobility signals. This modeling framework can be readily extended to regression problems with large scale spatio-temporal data -- in particular for our case, disease status reports and human mobility patterns at various temporal and geographical scales. In comparison to previous mechanistic or autoregressive approaches, our model does not rely on assumptions of the underlying disease dynamics and can learn from a variety of data, including inter-region interaction and region-level features.

There is still much to be done, both for \mbox{COVID-19} and for modeling infectious disease in general; we hope that this paper sparks an increased focus on leveraging this powerful new source of mobility information through novel techniques in graph learning. Future work can expand on these results by incorporating new features, expanding the time horizon for long term predictions, and experimenting on epidemiological mobility data in other parts of the world.

\bibliographystyle{ACM-Reference-Format}
\bibliography{paper}

\newpage

\section{Appendix}

\subsection{Google COVID-19 Aggregated Mobility Research Dataset}
\label{agg-epi-desc}
The Google COVID-19 Aggregated Mobility Research Dataset used for this study is available with permission from Google LLC. The Dataset contains anonymized mobility flows aggregated over users who have turned on the Location History setting, which is off by default. This is similar to the data used to show how busy certain types of places are in Google Maps --- helping identify when a local business tends to be the most crowded. The dataset aggregates flows of people from region to region, which is further aggregated at the level of US county, weekly in this study.

To produce this dataset, machine learning is applied to logs data to automatically segment it into semantic trips~\cite{bassolas2019hierarchical}. To provide strong privacy guarantees, all trips were anonymized and aggregated using a differentially private mechanism~\cite{wilson2020differentially} to aggregate flows over time\footnote{See https://policies.google.com/technologies/anonymization for more.}. This research is done on the resulting heavily aggregated and differentially private data. No individual user data was ever manually inspected, only heavily aggregated flows of large populations were handled.

All anonymized trips are processed in aggregate to extract their origin and destination location and time. For example, if  users traveled from location $a$ to location $b$ within time interval $t$, the corresponding cell $(a,b,t)$ in the tensor would be $n \pm err$, where $err$ is Laplacian noise. The automated Laplace mechanism adds random noise drawn from a zero mean Laplace distribution and yields $(\epsilon, \delta)$-differential privacy guarantee of $\epsilon=0.66$ and $\delta=2.1 \times 10^{-29}$ per metric. Specifically, for each week $W$ and each location pair $(A,B)$, we compute the number of unique users who took a trip from location $A$ to location $B$ during week $W$. To each of these metrics, we add Laplace noise from a zero-mean distribution of scale $\frac{1}{0.66}$. We then remove all metrics for which the noisy number of users is lower than 100, following the process described in https://research.google/pubs/pub48778/, and publish the rest. This yields that each metric we publish satisfies $(\epsilon,\gamma)$-differential privacy with values defined above. The parameter $\epsilon$ controls the noise intensity in terms of its variance, while $\gamma$ represents the deviation from pure $\epsilon$-privacy. The closer they are to zero, the stronger the privacy guarantees.

\end{document}